\documentclass[sigconf,natbib]{acmart}
\usepackage{multirow}
\usepackage{tcolorbox}
\usepackage{tabularx}
\usepackage{makecell}
\usepackage{siunitx}
\usepackage{booktabs}
\usepackage{float}
\usepackage{graphicx}
\usepackage{caption}
\usepackage{subcaption} % for subfigures
\usepackage{adjustbox}
\usepackage{enumitem}
\usepackage{arydshln}
\newcommand{\from}[3]{{\color{red}\small \textbf{\textsf{FROM #1 TO #2: #3}}}}

\AtBeginDocument{%
  \providecommand\BibTeX{{%
    \normalfont B\kern-0.5em{\scshape i\kern-0.25em b}\kern-0.8em\TeX}}}

\copyrightyear{2025}
\acmYear{2025}
\setcopyright{rightsretained}
\acmConference[SIGIR '25]{Proceedings of the 48th International ACM SIGIR Conference on Research and Development in Information Retrieval}{July 13--18, 2025}{Padua, Italy}
\acmBooktitle{Proceedings of the 48th International ACM SIGIR Conference on Research and Development in Information Retrieval (SIGIR '25), July 13--18, 2025, Padua, Italy}\acmDOI{10.1145/3726302.3730155}
\acmISBN{979-8-4007-1592-1/2025/07}

\begin{document}

%%
%% The "title" command has an optional parameter,
%% allowing the author to define a "short title" to be used in page headers.
\title{Combating Biomedical Misinformation through Multi-modal Claim Detection and Evidence-based Verification}

%%
%% The "author" command and its associated commands are used to define
%% the authors and their affiliations.
%% Of note is the shared affiliation of the first two authors, and the
%% "authornote" and "authornotemark" commands
%% used to denote shared contribution to the research.
\author{Mariano Barone}
\email{mariano.barone@studenti.unina.it}
\orcid{0009-0004-0744-2386}
\affiliation{%
  \institution{University of Naples Federico II}
  \city{Naples}
  \state{Italy}
  \country{Italy}
}

\author{Antonio Romano}
\email{antonio.romano5@unina.it}
\orcid{0009-0000-5377-5051}
\affiliation{%
  \institution{University of Naples Federico II}
  \city{Naples}
  \state{Italy}
  \country{Italy}
}

\author{Giuseppe Riccio}
\email{giuseppe.riccio3@unina.it}
\orcid{0009-0002-8613-1126}
\affiliation{%
  \institution{University of Naples Federico II}
  \city{Naples}
  \state{Italy}
  \country{Italy}
}

\author{Marco Postiglione}
\email{marco.postiglione@northwestern.edu}
\orcid{0000-0003-1470-8053}
\affiliation{%
  \institution{Northwestern University}
  \city{IL}
  \state{United States}
  \country{United States}
}

\author{Vincenzo Moscato}
\email{vincenzo.moscato@unina.it}
\orcid{0000-0002-0754-7696}
\affiliation{%
  \institution{University of Naples Federico II}
  \city{Naples}
  \state{Italy}
  \country{Italy}
}

%%
%% By default, the full list of authors will be used in the page
%% headers. Often, this list is too long, and will overlap
%% other information printed in the page headers. This command allows
%% the author to define a more concise list
%% of authors' names for this purpose.
\renewcommand{\shortauthors}{Barone et al.}

%%
%% The abstract is a short summary of the work to be presented in the
%% article.
\begin{abstract}
Biomedical misinformation --- ranging from misleading social media posts to fake news articles and deepfake videos --- is increasingly pervasive across digital platforms, posing significant risks to public health and clinical decision-making. We developed \textsf{CER}, a comprehensive fact-checking system designed specifically for biomedical content. CER integrates specialized components for claim detection, scientific evidence retrieval, and veracity assessment, leveraging both transformer models and large language models to process textual and multimedia content. Unlike existing fact-checking systems that focus solely on structured text, CER can analyze claims from diverse sources including videos and web content through automatic transcription and text extraction. The system interfaces with PubMed for evidence retrieval, employing both sparse and dense retrieval methods to gather relevant scientific literature. Our evaluations on standard benchmarks including HealthFC, BioASQ-7, and SciFact demonstrate that CER achieves state-of-the-art performance, with F1-score improvements compared to existing approaches. To ensure reproducibility and transparency, we release the GitHub repository with the source code, within which you can reach an interactive demonstration of the system published on HuggingFace and a video demonstration of the system \url{https://github.com/PRAISELab-PicusLab/CER-Fact-Checking}.
\end{abstract}

%%
%% The code below is generated by the tool at http://dl.acm.org/ccs.cfm.
%% Please copy and paste the code instead of the example below.
%%


%%
%% Keywords. The author(s) should pick words that accurately describe
%% the work being presented. Separate the keywords with commas.

%%
%% This command processes the author and affiliation and title
%% information and builds the first part of the formatted document.
\maketitle

\section{Introduction} \label{sec: introduction}

\begin{figure*}[t]
    \centering
    \includegraphics[width=\linewidth]{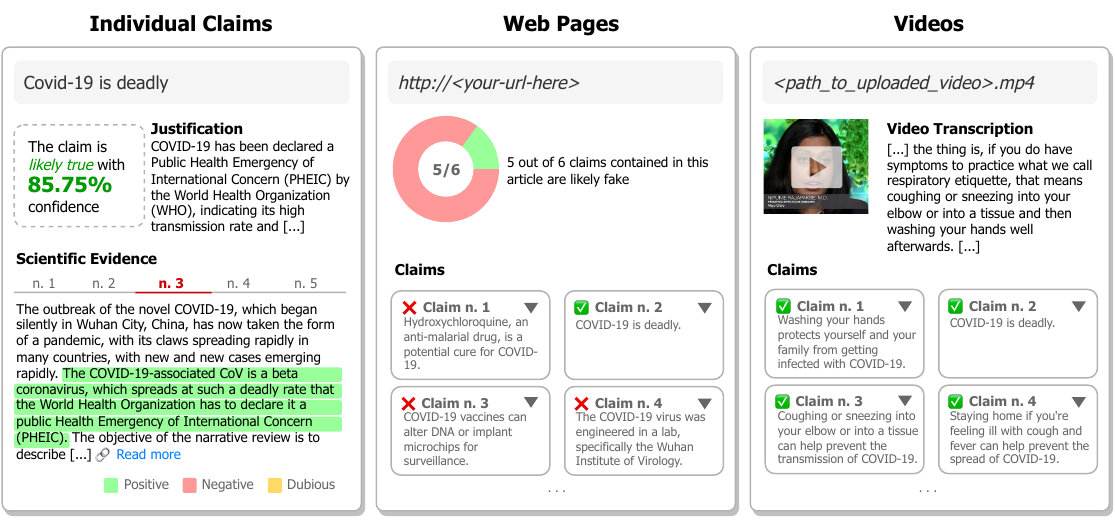}
    \caption{Examples showing system functionalities across three input types: individual claims (left), web pages (center), and video content (right). For claims derived from web pages and videos, users can expand the interface---as illustrated in the \textit{Individual Claims} panel---to examine claim justifications and scientific evidence that either supports or refutes each claim.}
    \label{fig:UI_example}
\end{figure*}

Biomedical misinformation threatens public health by influencing clinical decisions and eroding trust in medical institutions. Its spread can result in delayed treatments, higher mortality rates, and reliance on unverified practices \cite{johnson2022cancer,gisondi2022deadly}. Addressing this challenge requires advanced tools for claim verification. Recent AI-driven fact-checking developments have led to tools like factiverse.ai \cite{setty2024surprisingefficacyfinetunedtransformers}, designed to assess online information credibility. Other fact-checking systems, such as Sourcer AI\footnote{\url{https://sourcerai.co/}}  use machine learning models to evaluate the reliability of news articles and social media content. Another fact-checker application is FactCheck Editor \cite{setty2024factcheckeditormultilingualtext} that extends automated fact-checking capabilities to text editing workflows, supporting multilingual verification across over 90 languages. However, in biomedical and healthcare fields, fact-checking still relies heavily on manual verification, a time-consuming and complex process due to specialized terminology and the need for empirical evidence interpretation \cite{DBLP:journals/ijinfoman/Zhong23}. Moreover, existing automated solutions primarily focus on structured texts such as documents or datasets that follow a predefined format with clear organization, overlooking more diverse sources such as online publications and multimedia content, thereby limiting their effectiveness in countering misinformation across digital platforms. This issue is further exacerbated by the rise of deepfake technologies \cite{1282}, which enable the creation of fabricated yet highly realistic medical content \cite{202402.0176}, including altered videos of healthcare professionals, synthetic patient testimonies, and manipulated research presentations \cite{Patil.Shankargouda} . Such deceptive content can spread false medical advice or distort scientific consensus, increasing public confusion and skepticism toward legitimate healthcare recommendations \cite{DBLP:journals/corr/abs-2407-15169}. To overcome these limitations, this work introduces Combining Evidence and Reasoning (\texttt{CER}) %\from{Marco}{Mariano}{Abbiamo cambiato nome?} \answerfrom{mariano}{Marco}{Ops}, 
a multi-stage LLM-based system integrating claim detection, evidence retrieval, and veracity assessment. Unlike conventional approaches, \texttt{CER} processes both text and video content by extracting audio, generating transcriptions, and identifying claims for verification. This capability extends the reach of automated fact-checking to non-textual sources, enhancing its applicability in real-world misinformation scenarios. The system interfaces with PubMed for evidence retrieval, using dense retrieval to collect relevant scientific literature. \texttt{CER} is designed for researchers, fact-checkers, healthcare professionals, and policymakers, providing an effective tool for verifying biomedical claims with transparent, evidence-based explanations. It significantly reduces the efforts required to assess the credibility of claims while improving accuracy through advanced retrieval techniques and LLM-based reasoning \cite{DBLP:conf/coling/ZhangG24}. The system’s interface presents retrieved evidence alongside claim assessments, offering a clear and interpretable verification process. Experimental evaluations demonstrate that \texttt{CER} achieves state-of-the-art performance on benchmarks such as HealthFC \cite{DBLP:conf/coling/VladikaSM24}, BioASQ-7 \cite{DBLP:journals/corr/abs-2006-09174}, and SciFact \cite{wadden2020emnlp}, with F1-score improvements of up to +7.69\%. Furthermore, we demonstrate \textsf{CER}'s effectiveness as a complementary safeguard in deepfake detection pipelines, addressing a critical gap where traditional detectors struggle to generalize to emerging synthetic media formats amid rapidly evolving generation technologies \cite{DBLP:conf/iccv/YaoWD023}.

\begin{figure*}[ht]
    \centering \includegraphics[width=0.93\textwidth]{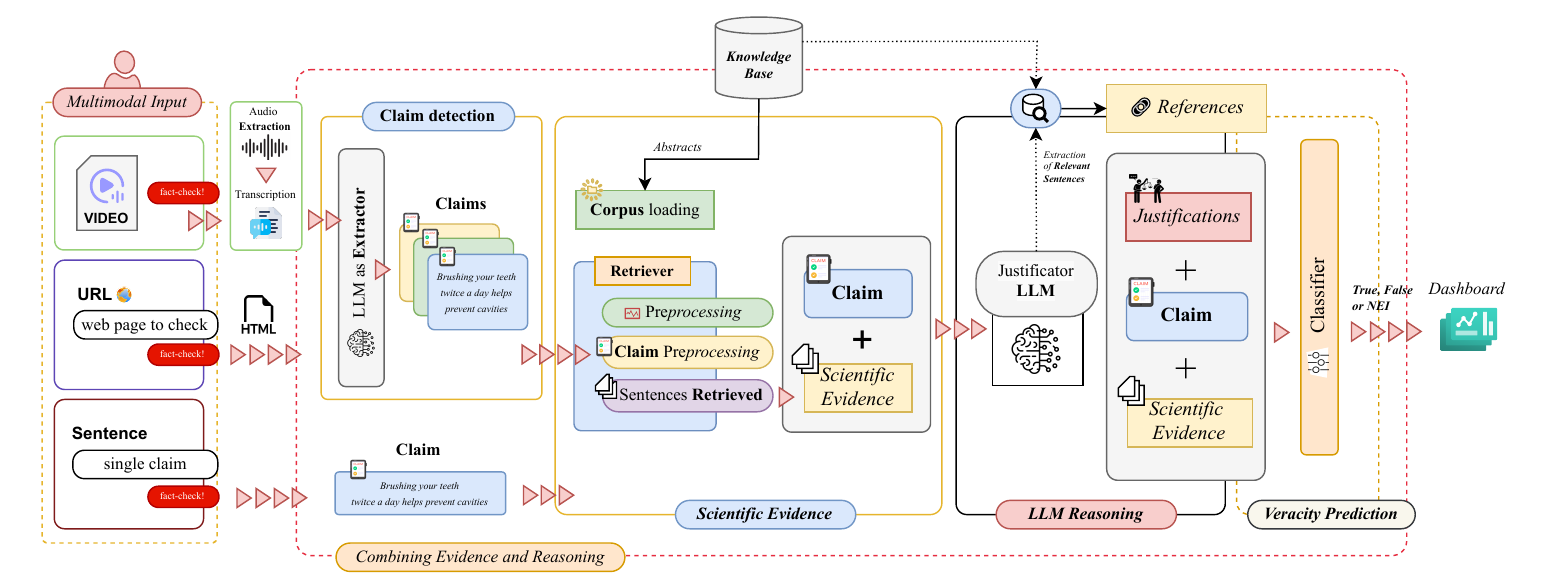}
    \caption{\textsf{CER} workflow. The system processes multimodal inputs, including text, URLs, and video content, by extracting and transcribing audio when necessary. Claims are detected using an LLM-based extractor and preprocessed before being matched against a scientific corpus indexed with a retrieval system. Relevant evidence passages are retrieved and combined with the original claim as input to an LLM, which generates detailed justifications. The system also incorporates a knowledge base for additional reference extraction. Finally, a binary classifier assesses the claim's veracity based on the claim text, retrieved scientific evidence, and generated justifications, producing a final classification (true/false/nei). The results are visualized through a dashboard for interpretability and decision support}
    \label{fig:CER_workflow}
    \Description{CER workflow}
\end{figure*}

\section{\texttt{CER}: Overview} \label{sec: methodology}

In this section, we present our proposed methodology to classify biomedical \textit{claims}\footnote{In this work, a \textit{claim} refers to any statement or question whose truthfulness or factual accuracy can be evaluated based on external evidence. This broad definition encompasses declarative sentences as well as interrogative forms.}.
Our architecture, as illustrated in Figure \ref{fig:CER_workflow}, is divided into four main components: (1) Claim Detection, (2) Scientific Evidence Retrieval, (3) LLM Reasoning and (4) Veracity Prediction. The remainder of this section provides an in-depth description of each component.

\begin{comment}
\begin{figure}[ht]
    \centering 
    \includegraphics[width=0.7\columnwidth]{img/Interfaccia/Interfaccia_colonna.png}
    \caption{\textsf{CER} - Biomedical Fact Checker demo interface. The system verifies the claim 'COVID-19 is deadly,' classifying it as 'Most likely True' with 85.75\% confidence. Supporting scientific evidence is highlighted in green, indicating positive evidence from relevant sources.}
    \label{fig:CER_interface}
    \Description{CER interface}
\end{figure}
\begin{figure}[ht]
    \centering 
    \includegraphics[width=0.7\columnwidth]{img/Interfaccia/Interfaccia_webpage.png}
    \caption{\textsf{CER} - Biomedical Fact Checker demo interface. The system extracts claims from a webpage and fact-checks each extracted claim independently}
    \label{fig:CER_interface2}
    \Description{CER interface_2}
\end{figure}
\begin{figure}[ht]
    \centering 
    \includegraphics[width=0.7\columnwidth]{img/Interfaccia/videoInterface.png}
    \caption{\textsf{CER} - Biomedical Fact Checker demo interface: The system transcribes dialogue from a video, extracts claims, and fact-checks each one.}
    \label{fig:CER_interface_3}
    \Description{CER interface_3}
\end{figure}
\end{comment}

\noindent \emph{ Inputs.}\;\textsf{CER} does not only handle plain text data, but also \textit{web pages} and \textit{videos}. When dealing with web pages, it extracts their HTML and identifies potential claims with an LLM-based method; the claim verification pipeline is then applied to all the claims. Similarly, audios are automatically extracted and transcribed when operating with videos, allowing the system to detect and validate claims from spoken content. \newline

\noindent \emph{ Claim Detection.}\; The claim detection phase leverages a LLM to identify statements worthy of verification within biomedical texts extracted from online sources, like articles and videos. The process begins with a text preprocessing step, including segmentation and tokenization, to prepare the input for LLM analysis. Claims are identified by classifying each sentence or paragraph based on its informational relevance, pertinence, and potential impact on clinical or public health decisions, using zero-shot or few-shot prompting techniques. When the input is a \textit{video}, the system first extracts the audio, then transcribes the content, and uses the LLM to spot verifiable claims. Transcription is performed using \textit{Whisper small-v3}, a robust multilingual speech recognition model capable of handling medical terminology and varied audio conditions. For \textit{web pages} (URLs), it analyzes the HTML code, extracting and processing the text to pinpoint pertinent claims. By incorporating LLMs in the claim detection step, \texttt{CER} guarantees scalability and flexibility in handling biomedical content from a variety of sources, such as textual documents, web pages, and multimedia. 
We use \textit{meta/llama-3.1-405b-instruct} for claim detection tasks, given its strong performance in instruction-following scenarios. %This model was prompted in a zero-shot ((or few-shot fashion)) to identify verifiable biomedical claims across modalities.
\newline

\noindent\emph{ Scientific Evidence Retrieval.} \; The process begins with a retrieval module interfacing with PubMed, a comprehensive repository of peer-reviewed biomedical literature. Focusing on abstracts, which offer dense and accessible research summaries, evidence is extracted using Dense Retrieval (multi-qa-MiniLM-L6-cos-v1) with a FAISS index. Dense Retrieval employs SBERT embeddings to capture semantic nuances. 
%Although SBERT is not specialized for biomedical language, we mitigate terminology mismatch by applying biomedical-aware text normalization and stopword filtering. Retrieval quality was also manually validated on a random sample. 
Future extensions may incorporate BioBERT-based retrieval for better domain coverage. Preprocessing includes normalization, tokenization, and stopword removal. Claims are processed as queries, and the top 20 results are ranked for relevance. For each claim, up to three evidence pieces are selected and formatted as a concatenated structure with a separator token (\texttt{[SEP]}), balancing information coverage and computational efficiency. These structured claim-evidence pairs are then prepared for reasoning. \newline

\noindent \emph{LLM Reasoning.}\; The reasoning phase employs large language models (LLMs) as assistants to assess claim veracity and generate justifications, mitigating the risk of hallucinations. We use \textit{meta-llama/Meta-Llama-3.1-405B-Instruct}, an instruction-tuned language model by Meta, chosen for its strong reasoning capabilities and coherent, context-aware justifications. Its robustness across biomedical domains makes it ideal for tasks requiring factual accuracy and explanatory output. Given a structured prompt with claim and evidence, the model returns a binary veracity assessment and a detailed justification. %The prompt explicitly instructs the model to base its reasoning only on the provided evidence, discouraging external assumptions. Additionally, decoding parameters (e.g., low temperature and repetition penalties) are tuned to favor grounded, factual outputs. 
The outputs are passed to the subsequent veracity prediction stage, ensuring the reasoning aligns with retrieved evidence. This design ensures transparency and accuracy in evaluating biomedical claims. 
Notably, while the LLM provides a preliminary binary veracity judgment, this output is not directly used in the final classification. Instead, the justification is combined with the original claim and fed into the classification model (e.g., DeBERTa or BERT) to determine the final label.
\newline

\noindent \emph{Veracity Prediction.}\; The final stage integrates the outputs of the LLM reasoning module into a classification framework to assign one of three labels: "true," "false," or "nei (not enough information)." Two approaches are explored: zero-shot classification and fine-tuning. The zero-shot approach leverages pre-trained models to classify claims without task-specific training, providing rapid deployment but limited domain precision. Fine-tuning adapts the model using labeled domain-specific data, optimizing accuracy and relevance at the cost of additional computational resources. This stage ensures robust and reliable fact-checking tailored to biomedical applications.  %The classifier receives both the claim and the LLM-generated justification as input, leveraging the justification to improve label assignment. This architecture allows us to assess how explanations produced by the LLM contribute to veracity classification.

\begin{comment}
\section{Inputs}
\from{Marco}{Mariano}{Questa sezione va integrata in Sezione 2. Va aggiunto un punto prima di \textit{Claim Detection}}
\textsf{CER} stands out for its high versatility in processing data from heterogeneous sources, dynamically adapting to different input types. The system can handle:  

\begin{itemize}
    \item Plain text: claims are directly analyzed through the claim detection and verification pipeline.  
    \item Web pages (URLs): the HTML is extracted and processed by an LLM, which identifies potentially verifiable claims; the claim verification pipeline is then applied to the relevant text.
    \item Videos: the audio is extracted and transcribed, allowing \textsf{CER} to detect and validate claims from spoken content before applying the claim detection and verification pipeline.  
\end{itemize}

This \textit{operational flexibility} enables \textsf{CER} to adapt to complex real-world scenarios, ensuring effective claim analysis regardless of the input format.  
\end{comment}

\begin{table*}[t]
  \centering
  \begin{adjustbox}{width=\textwidth} % Scala la tabella per adattarla alla larghezza della pagina
    \begin{tabular}{llccccccccccc}
      \toprule
      \textbf{Category} & \textbf{Fact-Checker} & \multicolumn{3}{c}{\textbf{HealthFC}} & & \multicolumn{3}{c}{\textbf{BioASQ}} & & \multicolumn{3}{c}{\textbf{SciFact}} \\
      \cline{3-5} \cline{7-9} \cline{11-13}
                       &                         & \textbf{Precision (\%)} & \textbf{Recall (\%)} & \textbf{F1 (\%)} & & \textbf{Precision (\%)} & \textbf{Recall (\%)} & \textbf{F1 (\%)} & & \textbf{Precision (\%)} & \textbf{Recall (\%)} & \textbf{F1 (\%)} \\
      \midrule
      Baselines         & All True (baseline)     & 8.97 & 33.33 & 14.14 & & 82.39 & 1.0 & 90.34 & & 13.71 & 33.33 & 19.43 \\
                       & All False (baseline)    & 5.55 & 33.33 & 9.52  & & 17.60 & 1.0 & 29.94 & & 7.16  & 33.33 & 11.78 \\
                       & All NEI (baseline)      & 18.79 & 33.33 & 24.04 & & -- & -- & -- & & 12.47 & 33.33 & 18.15 \\ \hline
      Online platforms        & Factiverse.AI \cite{DBLP:conf/sigir/Setty24a} & 53.47 & 30.19 & 38.59 & & 91.31 & 84.01 & 87.51 & & 50.25 & 46.80 & 48.46 \\
                       & Perplexity.AI\footnotemark[8] & 56.40 & 47.20 & 51.39 & & 83.20 & 80.13 & 81.64 & & 52.75 & 50.00 & 51.34 \\ \hline
      LLM Predictions  & Mixtral LLM prediction\footnotemark[10] & 40.10 & 41.24 & 40.61 & & 72.43 & 70.69  & 71.54 & & 45.89 & 40.60 & 43.02 \\
                       & GPT4o mini             & 40.52 & 39.88 & 39.06  & &  53.85 & 52.71 & 52.79& & 39.01 & 37.64 & 38.31 \\
                        \hline
                    
      Literature baselines     
      & \citet{DBLP:journals/corr/abs-2404-08359} & 52.60 & 44.50 & 40.60 & & 60.90 & 64.40 & 61.70 & & 46.00 & 42.30	 & 44.07 \\
      & \citet{10.1145/3485127} & 41.72 & 49.36 & 45.21 & & 47.42 & 52.36 & 49.76 & & 54.12 & 51.20 & 52.62 \\
                       & \citet{zaheer2020nips} & 26.67 & 35.95 & 30.62 & & 69.62 & 33.57 & 45.29 & & 38.70 & 35.20 & 36.87 \\
                       & \citet{DBLP:journals/corr/abs-1909-11942} & 21.40 & 39.36 & 27.72 & & 74.60 & 50.37 & 60.13 & & 34.45 & 31.85 & 33.10 \\
                       & \citet{DBLP:conf/coling/VladikaSM24} & 68.24 & 66.84 & 67.53 & & - & - & - & & - & - & - \\ \hline
                       
                    CER (\textit{zero-shot})   & \textsf{CER} (DeBERTA v3-large) & 58.26 & 51.48 & 54.61 & & 93.10 & 90.48 & 91.77 & & 55.72 & 51.90 & 53.74
                        \\ \hdashline
                       
      CER (\textit{fine-tuned})              & \textsf{CER} (DeBERTA v3-large) & 64.55 & 70.14 & 67.22 & & \textbf{93.70} & \textbf{96.75} & \textbf{95.20} & & \textbf{61.44} & \textbf{60.72} & \textbf{61.14} \\
                       & \textsf{CER} (PubMedBERT) & 67.55 & \textbf{72.43} & \textbf{69.90} & & 89.76 & 90.02 & 89.88 & & 57.70 & 56.13 & 56.91 \\
                       & \textsf{CER} (BERT)       & \textbf{69.92} & 69.33 & 69.62 & & 81.28 & 83.89 & 82.56 & & 58.10 & 55.89 & 56.97 \\
      \bottomrule
    \end{tabular}
  \end{adjustbox}
  \caption{\textbf{Comparison with state-of-the-art baselines}. This table presents the macro precision, recall, and F1 scores for different baseline models evaluated on the experimented datasets. Best results are highlighted in bold. '--' indicates instances where a technique could not be applied. }
  \label{tab:results_healthfc_bioasq}
\end{table*}

\section{Experiments} \label{sec: experiments}

This section presents an empirical evaluation of the \textsf{CER} architecture across multiple datasets and configurations. Our evaluation focuses on comparing our proposed approach with both fact-checking and deepfake detection baselines.

\subsection{Experimental Setup}

\subsubsection{Datasets}
We evaluate \textsf{CER} on three established fact-checking benchmark datasets and a novel video dataset. For text-based evaluation, we use HealthFC \cite{DBLP:conf/coling/VladikaSM24}, which contains 750 health-related claims verified by medical professionals and labeled as true, false, or NEI (Not Enough Information); BioASQ-7b \cite{DBLP:journals/corr/abs-2006-09174}, comprising 745 biomedical claims labeled as true or false; and SciFact \cite{wadden2020emnlp}, consisting of 1.4k scientific claims labeled as "support", "confute", or "NEI", with accompanying expert-written rationales for validation. To assess \textsf{CER}'s effectiveness on deepfake content, we created a balanced dataset of 40 healthcare videos: 20 deepfake videos collected from public sources and 20 authentic videos containing verified healthcare information\footnote{Given the limited size of our benchmark dataset, the deepfake detection results should be considered preliminary.}.

\begin{comment}
\subsubsection{Implementation and Training details}
The Dense Retriever was implemented on a system with an Intel i7-9700K CPU, 32GB RAM, and a GT710 GPU. Training was conducted on a machine with an Intel Xeon CPU, 13GB RAM, and an NVIDIA Tesla T4 GPU. The LLM for the "LLM Reasoning" module is \textit{Mixtral-8x22B-Instruct-v0.1} \cite{jiang2024mixtralexperts}, a Sparse Mixture of Experts (MoE) model suitable for domain-specific tasks. For the "Veracity Prediction" module, we tested DeBERTa \cite{he2023iclr}, BERT \cite{DBLP:journals/corr/abs-1810-04805}, and PubMedBERT \cite{gu2022health}.

Veracity Prediction models were trained using the original training and validation datasets, with hyperparameters optimized through grid and random search to ensure generalization and prevent overfitting. Performance was assessed on an independent test set (as provided by dataset authors) to ensure unbiased comparisons. Evaluation was based on macro precision, macro recall, and macro F1 scores.
\end{comment}

\subsubsection{Baselines}
We compared \textsf{CER} with open- and closed-source fact-checkers. Factiverse.AI and Perplexity.AI are closed-source platforms that classify evidence as "Supporting", "Mixed", or "Disputing" using sources like Google, Bing, Wikipedia, and FactSearch. \citet{DBLP:journals/corr/abs-2404-08359} propose a voting-based architecture similar to ours, while \citet{10.1145/3485127} focus on Wikipedia document and sentence retrieval for claim validation. Other methods include BigBird \cite{zaheer2020nips}, ALBERT \cite{DBLP:journals/corr/abs-1909-11942}, and \citet{DBLP:conf/coling/VladikaSM24}'s pipeline for medical claim fact-checking. Simple baselines ("All True", "All False", "All NEI") were also included. Additionally, we compared our proposed method with several deepfake detection baselines: Face X-Ray \cite{li2020face}, ResNet-34 \cite{wang2020cnn}, and F3Net \cite{qian2020thinking}.

\begin{table}[t]
  \centering
  \begin{adjustbox}{width=0.75\linewidth} % Scala la tabella per adattarla alla larghezza della pagina
    \begin{tabular}{llccccccccccc}
      \toprule
      \textbf{Method} & \multicolumn{3}{c}{\textbf{\textit{Real} class}} & \multicolumn{3}{c}{\textbf{\textit{Fake} class}} \\ \cline{2-4} \cline{5-7}
       & \textbf{P} & \textbf{R} & \textbf{F1} & \textbf{P} & \textbf{R} & \textbf{F1} \\ \midrule

       Face X-Ray & 1.00 & 1.00 & 1.00 & 1.00 & 0.90 & 0.95 \\
       ResNet-34 & 1.00 & 0.89 & 0.94 & 1.00 & 0.80 & 0.89 \\
       F3Net & 1.00 & 0.89 & 0.94 & 1.00 & 0.45 & 0.62 \\ \midrule
       CER & 1.00 & 1.00 & 1.00 & 1.00 & 1.00 & 1.00 \\
      \bottomrule
    \end{tabular}
  \end{adjustbox}
  \caption{\textbf{Comparison with deepfake detection baselines}. This table presents the precision (P), recall (R), and F1 scores for different baseline models and the two classes (i.e. real, fake).}
  \label{tab:results_video}
\end{table}

\subsection{Results} \label{sec: results}
We evaluated \textsf{CER} against state-of-the-art fact-checking baselines using both general-domain models (BERT, DeBERTa-v3) and a domain-specific biomedical model (PubMedBERT) in zero-shot and fine-tuned configurations. As shown in Table \ref{tab:results_healthfc_bioasq}, fine-tuned \textsf{CER} consistently outperforms all baselines across datasets, achieving F1-scores of 69.90\% on HealthFC, 95.20\% on BioASQ, and 61.14\% on SciFact. The model maintains competitive performance even in zero-shot settings, reaching 54.61\% on HealthFC, 91.77\% on BioASQ, and 53.74\% on SciFact. While PubMedBERT leverages domain-specific biomedical pretraining, it achieves lower performance on both SciFact (56.91\%) and BioASQ (89.88\%). This performance gap highlights the advantages of combining general-domain knowledge with task-specific fine-tuning. Similarly, although DeBERTa-v3 demonstrates strong results on BioASQ (95.20\%) and SciFact (61.14\%), it falls short of \textsf{CER}'s performance on HealthFC and BioASQ. Table \ref{tab:results_video} presents a comparison between \textsf{CER} and established deepfake detection baselines. Our classification strategy designated a video as "fake" if our system identified at least one false claim within its content. Interestingly, this approach achieved perfect accuracy across the entire benchmark dataset, surpassing the performance of all evaluated deepfake detectors. While these results are preliminary, they highlight \textsf{CER}'s potential as a complementary component in deepfake detection pipelines. This is especially significant given that traditional deepfake detectors often struggle to generalize to novel forms of synthetic media, due to the rapid evolution of deepfake generation technologies \cite{DBLP:conf/iccv/YaoWD023}.

\section{Conclusion} \label{sec: conclusion}

We introduced \textsf{CER} (Combining Evidence and Reasoning), a biomedical fact-checking system with a user-friendly demo. It verifies medical claims, web pages, and videos by integrating scientific evidence retrieval with LLM-based reasoning. This structured approach balances factual accuracy and interpretability, offering explainable responses with accurate sources.

\begin{comment}
Future work should expand evidence retrieval beyond current biomedical sources to include clinical trial registries and real-time medical updates.
\end{comment}

\begin{comment}
\paragraph{Limitations} The architecture relies heavily on pre-trained language models for the generation of justifications. These models have known limitations, including a tendency to generate hallucinated content (incorrect but plausible-sounding information) \cite{DBLP:journals/corr/abs-2310-01469}, particularly when dealing with zero-shot tasks. While our pipeline mitigates this issue with the supervised \textit{Veracity Prediction} step, generating justifications with models that are free from the hallucinations problem could potentially improve prediction accuracy. 
\end{comment}

% \section*{Ethical Considerations}
% This paper implements a fact-checking system for medical claims. Although promising results, they are not yet ready for everyday use due to incorrect predictions caused by hallucinations, lack of information, and difficulty in interpreting models, which could lead to harmful consequences and further spread misinformation.

\section*{Acknowledgements}

This work was conducted with the financial support of (1) the PNRR MUR project PE0000013-FAIR and (2) the Italian ministry of economic development, via the ICARUS (Intelligent Contract Automation for Rethinking User Services) project (CUP: B69J23000270005).

\begin{comment}
\begin{figure}[ht]
    \centering
    \begin{subfigure}{0.6\textwidth} 
        \centering
        \includegraphics[width=1\textwidth]{img/Appendix/2.pdf}
        \caption{Example from the HealthFC dataset with generated justification and label predicted using fine-tuning.}
        \label{fig:NEI_example}
    \end{subfigure}
    \hfill
    \begin{subfigure}{0.6\textwidth} 
        \centering
        \includegraphics[width=1\textwidth]{img/Appendix/3.pdf}
        \caption{Example from the BioASQ dataset with generated justification and label predicted using fine-tuning.}
        \label{fig:Yes_example}
    \end{subfigure}
    \caption{Examples from the HealthFC and BioASQ datasets with generated justifications and labels predicted using fine-tuning.}
    \label{fig:CER_Examples}
    \Description{CER examples}
\end{figure}
\end{comment}

\newpage
%% The next two lines define the bibliography style to be used, and the bibliography file.
\bibliographystyle{ACM-Reference-Format}
\bibliography{sample-base}

\end{document}